\documentclass[letterpaper, 10 pt, conference]{ieeeconf}

\IEEEoverridecommandlockouts
\usepackage{amsmath} 
\usepackage{amssymb}  
\usepackage[protrusion=false]{microtype}
\usepackage{tensor}
\usepackage{mathtools}
\usepackage{makecell}
\usepackage[colorinlistoftodos,prependcaption]{todonotes}
\usepackage{siunitx}
\usepackage{comment}
\usepackage{xcolor} 
\usepackage[normalem]{ulem}
\usepackage{esvect}

\usepackage{booktabs}

\title{\LARGE \bf
	Extrinsic Calibration and Verification of Multiple Non-overlapping Field of View Lidar Sensors}

\author{Sandipan Das$^{1,2}$, Navid Mahabadi$^{2}$, Addi Djikic$^{2}$, Cesar Nassir$^{2}$, Saikat Chatterjee$^{1}$, Maurice Fallon$^3$
	\thanks{$^1$ KTH EECS, Sweden. \texttt{\{sandipan,sach\}@kth.se}\newline%
		$^2$ Scania CV AB, Sweden. \texttt{\{sandipan.das,navid.mahabadi,addi. djikic,cesar.nassir\}@scania.com}\newline%
		$^3$Oxford Robotics Institute, UK. \texttt{mfallon@robots.ox.ac.uk}}}

\usepackage[linesnumbered,ruled,vlined]{algorithm2e}

\usepackage{multirow}
\usepackage[normalem]{ulem}
\usepackage{xargs} 

\newcommand{\hide}[1]{}
\newcommand{\Figure}{Fig.~}

\newcommand{\bdmath}{\begin{dmath}}
\newcommand{\edmath}{\end{dmath}}
\newcommand{\beq}{\begin{equation}}
\newcommand{\eeq}{\end{equation}}
\newcommand{\bdm}{\begin{displaymath}}
\newcommand{\edm}{\end{displaymath}}
\newcommand{\bea}{\begin{eqnarray}}
\newcommand{\eea}{\end{eqnarray}}
\newcommand{\beal}{\beq \begin{array}{ll}}
\newcommand{\eeal}{\end{array} \eeq}
\newcommand{\beas}{\begin{eqnarray*}}
\newcommand{\eeas}{\end{eqnarray*}}
\newcommand{\ba}{\begin{array}}
\newcommand{\ea}{\end{array}}
\newcommand{\bit}{\begin{itemize}}
\newcommand{\eit}{\end{itemize}}
\newcommand{\ben}{\begin{enumerate}}
\newcommand{\een}{\end{enumerate}}

\newcommand{\SO}{\mathrm{SO}}

\newcommand{\Real}{\mathbb{R}}

\newcommand{\SEthree}{\ensuremath{\mathrm{SE}(3)}\xspace}

\newcommand{\SOthree}{\ensuremath{\SO(3)}\xspace}

\newcommand{\calK}{{\cal K}}

\newcommand{\calN}{{\cal N}}

\newcommand{\calP}{{\cal P}}

\newcommand{\calT}{{\cal T}}

\newcommand{\T}{\mathbf{T}}

\newcommand{\eye}{{\mathbf I}}

\newcommand{\Ithree}{\eye_{3 \times 3}}

\newcommand{\World}{\mathtt{W}}
\newcommand{\Imu}{\mathtt{I}}
\newcommand{\Camera}{\mathtt{C}}
\newcommand{\Lidar}{\mathtt{L}}

\newcommand{\Base}{\mathtt{{B}}}

\SetCommentSty{algo}
\SetKwInput{KwInput}{Input} 
\SetKwInput{KwOutput}{Output} 

\DeclareMathOperator*{\argmax}{arg\,max}
\DeclareMathOperator*{\argmin}{arg\,min}

\newcommand\cancel{\bgroup\markoverwith{\textcolor{red}{\rule[0.5ex]{2pt}{0.4pt}}}\ULon}

\makeatletter
\let\NAT@parse\undefined
\makeatother
\usepackage{hyperref}

\begin{document}
	
	\maketitle
	\thispagestyle{empty}
	\pagestyle{empty}
	
	\begin{abstract}
		We demonstrate a multi-lidar calibration framework for large mobile platforms that jointly calibrate the extrinsic parameters of non-overlapping Field-of-View (FoV) lidar sensors, without the need for any external calibration aid. The method starts by estimating the pose of each lidar in its corresponding sensor frame in between subsequent timestamps. Since the pose estimates from the lidars are not necessarily synchronous, we first align the poses using a Dual Quaternion (DQ) based Screw Linear Interpolation. Afterward, a Hand-Eye based calibration problem is solved using the DQ-based formulation to recover the extrinsics. Furthermore, we verify the extrinsics by matching chosen lidar semantic features, obtained by projecting the lidar data into the camera perspective after time alignment using vehicle kinematics. Experimental results on the data collected from a Scania vehicle [$\sim$ 1 Km sequence] demonstrate the ability of our approach to obtain better calibration parameters than the provided vehicle CAD model calibration parameters. This setup can also be scaled to any combination of multiple lidars. 
	\end{abstract}
	
	
	

	\section{Introduction}
	\label{sec:introduction}
\begin{figure}
		\centering
		\includegraphics[width=1\columnwidth]{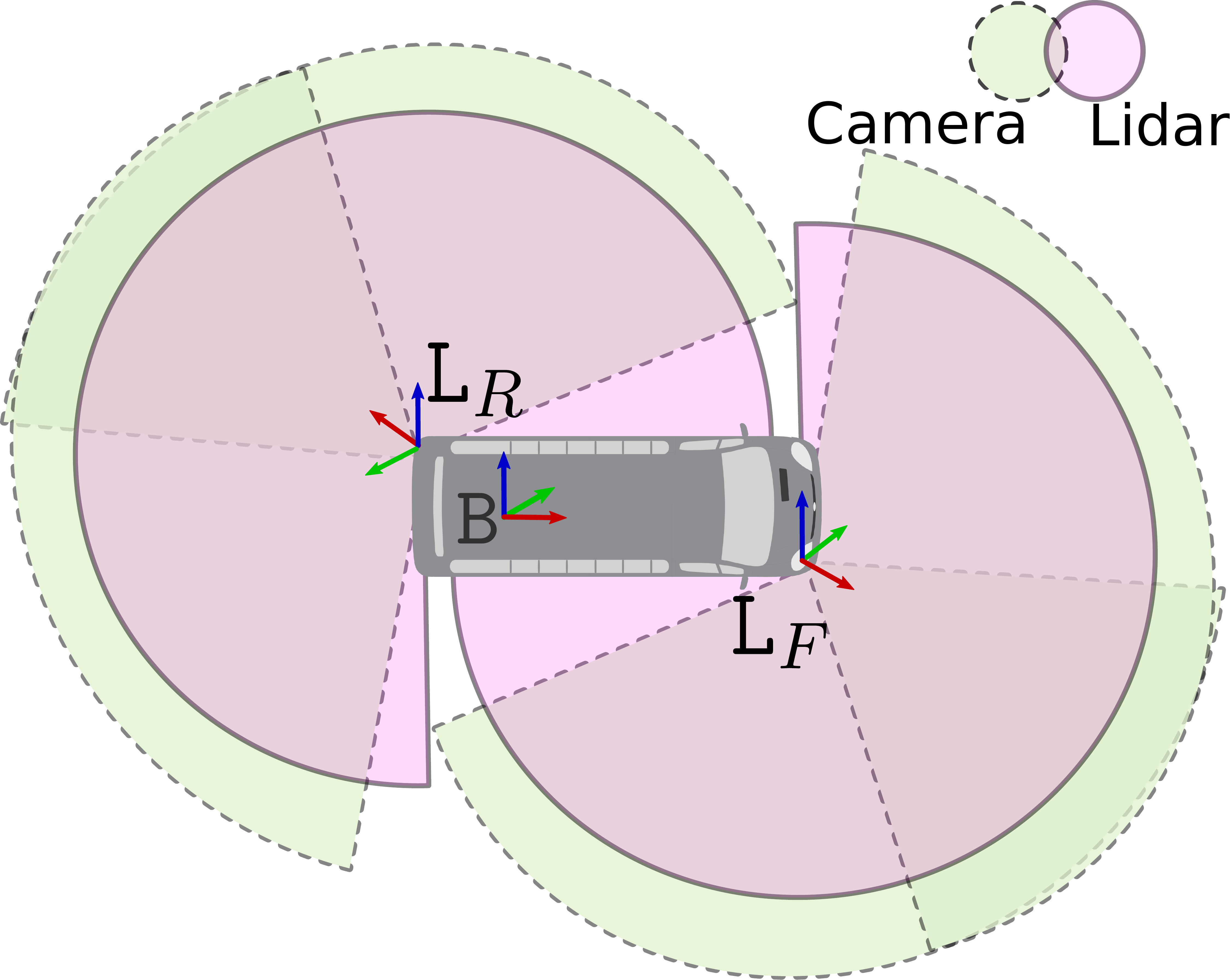}
		\caption{Illustration of the Field of View (FoV) of two lidars and four cameras positioned around the data collection vehicle. The scale factors of the sensor ranges and the vehicle dimension are only for illustrative purposes. The vehicle base frame $\Base$, is located at the center of the rear axle. The sensor frames of the 2 lidars are: $\Lidar_{{F}|{R}}$, where $F$ and $R$ are the front right and rear left lidars.} 
		\label{fig:scania-bevda}
		\vspace{-5mm}
	\end{figure}

	Multi-sensor calibration is one of the most important prerequisites for successful mobile autonomy in the real-world. To have complete $360^{\circ}$ sensing coverage, mobile vehicles often have multiple sensors around the vehicle, and in fact there can be extra sensors for redundancy and safety reasons (see \Figure\ref{fig:scania-bevda}). This complexity poses a challenge when calibrating the sensors to an acceptable level before perceiving the environment coherently.      
	
	Recent advances in lidar sensing have promoted their usage in the sensing stack of autonomous vehicles including wider Field-of-View (FoV), longer range and higher resolution. Since, lidar sensors can provide range information along with semantics
	\cite{pointnet_plus_plus_qi_2017} they have become a key component in the perception pipeline of autonomous platforms.   
	
	However, significant calibration degradation can occur due to issues such as thermal expansion/contraction of the lidar mount relative to the platform; navigating on rough terrain (as found in a mine) or the general effect of wear and tear over the long run. Hence, it is imperative to be able to re-calibrate, adopt and update the lidar extrinsic parameters \textit{in situ}. Offline calibration or target-based lidar calibration processes \cite{Huang2009, Pandey2010, Gao2010, Choi2016, Zhou2018, Kim2020} can be time-intensive and costly for large mobile platforms. Marker-less extrinsic calibration with state optimization \cite{Kuemmerle2011b}  or
	trajectory optimization \cite{Tsai_Hand_Eye_1989, Taylor_2015, Taylor_2016} are more desirable because of their time efficiency and ease of usage by an end-user.

	\subsection{Motivation}
	
	To create an efficient extrinsic calibration framework suitable for multiple lidars we started with data collected using a Scania bus with similar sensor FoVs to that shown in Fig. \ref{fig:scania-bevda}. In spite of the various available solutions for multi-lidar calibration there were two major challenges. Problem 1) Although the sensors were synchronized with PTP synchronization \cite{vallat2007clock}, the estimated poses from the state estimation which were used for motion based calibration were not necessarily time synchronized. Problem 2) Because of the limited FoV overlap between the front right and the rear left lidar, we could not verify the quality of the estimated calibration as it was not possible to inspect if the point clouds expressed a consistent view of the surroundings.
	
	Prior work has addressed the time alignment issue by correlating the angular velocity norms of both pose signals \cite{furrer2018evaluation}. Further calibration refinement has been performed based on overlapping FoV \cite{Jiao2019, Jiao2021}. For non-overlapping FoV scenario, calibration verification was handled by matching features in a global map \cite{Carrera2011} in the visual domain.
	
	In our work, to address Problem 1) we have used dual quaternion (DQ) \cite{daniilidis1996dual} based time alignment of the poses using screw linear interpolation (ScLERP) \cite{kavan2005spherical} as it provides the shortest path between pose estimates on $\SEthree$ manifold. We have tackled Problem 2) by evaluating the correspondences of selected similar features observed from non-overlapping FoV lidars at different timestamps, analyzed based on vehicle kinematics. We have additionally used cameras, which were placed below the corresponding lidars as shown in Fig. \ref{fig:coordinate-frames} to improve the quality of feature selection from point clouds. 
	
	
	\subsection{Contribution}
	\label{sec:contribution}
	
	
	Our work is motivated by the Hand-Eye  calibration method of Tsai \textit{et al}~\cite{Tsai_Hand_Eye_1989}. Our specific contributions are additional procedures which increase the accuracy and robustness of the extrinsic calibration process as follows:
	\begin{itemize}
		\item A DQ based prior time alignment of the poses based on ScLERP before applying the DQ based Hand-Eye calibration.
		\item An additional verification method using semantic matching to evaluate the extrinsic calibration quality for non-overlapping FoV lidars.
		\item Experimental results and verification using data collected with a Scania bus mounted with the sensor setup as shown in Fig. \ref{fig:coordinate-frames}, with FoV schematics similar to Fig. \ref{fig:scania-bevda}.
	\end{itemize}
	
	
	
	\section{Related Work}
	\label{sec:related-work}
	
	There are several techniques to obtain the extrinsic calibration parameters of the multi-sensor platform. This section categorizes those methods and briefly describes the relevant work	in each category.
	
	\subsection{SLAM-based methods}
	Simultaneous Localization and Mapping (SLAM) approaches normally estimate the extrinsic parameters by considering it as a state variable in the pose graph optimization or bundle adjustment optimization in visual SLAM. Kummerle \textit{et al}~\cite{Kuemmerle2011b} extended the SLAM hyper graph formulation to handle calibration and kinematic parameters without prior knowledge of the environment by including wheels radii and sensor position as state variables. 
	Jones \textit{et al}~\cite{Jones2011}, proposed a visual inertial SLAM system that localized the sensor positions in the map frame to estimate the calibration parameter between a camera and an inertial measurement unit (IMU). This approach relied heavily on a robust SLAM solver along with a rich motion sequence. Similarly, Carrera \textit{et al}~\cite{Carrera2011} presented a visual SLAM algorithm for multi-camera extrinsic calibration by building a global landmark map from each camera and then fusing the maps by means of feature correspondences using similarity transformation bundle adjustment. 
	This approach required driving in a circular path as the basic prerequisite for calibration to ensure the observability of similar features. 
	To address this, Heng \textit{et al}~\cite{Heng2013} extended the prior work by incorporating wheel odometry and formulating a least-squares problem. Their method was restricted to ground robots with odometry. In contrast, their recent work \cite{Heng2015} had no odometry requirement and utilized visual SLAM and bundle adjustment using a multi-camera system where each sensor built its own map. Map merging was then performed using a hand-eye calibration method and the results were utilized in pose graph optimization to estimate the extrinsics among sensors. 
	
	
	\subsection{Appearance-based methods}
	The recent appearance based approaches can be clustered into two categories; first, those using artificial markers or a known calibration target (e.g. a checkerboard), and second automatic geometric feature extraction methods which do not rely on markers. Appearance based methods in general could be seen as registration problems where artificial markers or geometrical features are employed. Gao \textit{et al}~\cite{Gao2010}, for instance, recovered the calibration parameters between multiple lidars by manually observing reflective landmarks in the scene. 
	Zhou \textit{et al}~\cite{Zhou2018}, presented a method to calibrate lidar and camera sensor modalities using a checkerboard by combining 3D line and plane features. Similarly, in \cite{Huang2009}, \cite{Pandey2010} and \cite{Kim2020}, the authors proposed lidar to camera calibration by utilizing plane feature registration using a chessboard. Automatic data extraction from sensors makes the usage of artificial markers redundant. Choi et al. \cite{Choi2016} determined the relative extrinsic parameters between 2D-lidars by extracting orthogonal planes and utilizing the co-planar features of the scan points. Geometric features such as points, lines and planes were extracted in \cite{He2013} and matched to calibrate a dual lidar system. Levinson \textit{et al}~\cite{Levinson2013} presented an online camera to lidar calibration optimizer by aligning edge points from camera images and corresponding lidar points. However, for all these methods to work the general assumption was that the features were visible through an overlapping FoV correspondence.

	\subsection{Motion-based methods}
	Incremental motion based approaches recover the calibration parameters using the well-known hand-eye calibration formulation, \cite{Tsai1989}. The calibration was estimated between a gripper (hand) and a camera (eye) by moving the gripper in some trajectory, while exploiting the fact that the transformation between the camera and the gripper was always the same. More recent research has focused on extending the work to multi-sensor modalities by aligning the motion of each individual sensor. Brookshire \textit{et al}~\cite{Brookshire2013} utilized the DQ method \cite{Daniilidis1999} to recover the relative transformation of two camera sensors mounted on a rigid body. The same principle was also applied by Heng \textit{et al}~ in camodocal \cite{Heng2013}. A more generic sensor configuration was proposed by Taylor et al. \cite{Taylor_2015} for multi-modal sensor arrays calibration including 3D-lidar, IMU and camera by motion combination of sensors using hand-eye formulation. They also provided the uncertainty of the calibration. The same authors extended the work to incorporate the sensor's accuracy reading and estimated the timing offsets between the sensors as well \cite{Taylor_2016}. Moreover, overlapping fields of view was applied to further enhance the calibration when applicable. Jiao \textit{et al}~\cite{Jiao2019} exploited the hand-eye calibration of a multi-lidar system by motion alignment to recover an initial estimation of the extrinsic. A refinement process using point-to-plane features was then used to further enhance the	calibration parameters. In their recent work M-LOAM \cite{Jiao2021}, they used similar techniques with edge and planar features for the calibration refinements.
	
	Both SLAM based approach and motion based approach can be used to solve the extrinsic calibration problem for our system. In our work we adapted a motion based method to solve the extrinsic calibration problem due to the accuracy of the poses estimated via LOAM \cite{zhang2014loam} based approaches. We performed the trajectory alignment as an initialization step similar to \cite{Jiao2019}. With this initialization, we first did the time alignment based on DQ based ScLERP. After that we recovered the individual sensor poses and formulated the problem as a Hand-Eye calibration problem. The Hand-Eye problem was solved using a DQ based closed form solution \cite{Daniilidis1999}. Finally, we developed a verification method for evaluating the quality of the calibration by semantic matching of the chosen lidar features.
	

	
	\section{Problem Statement}
	\label{sec:problem-statement}
	
	\subsection{Reference Frames}
	\begin{figure}
		\centering
		\vspace{2mm}
		\includegraphics[width=1\columnwidth]{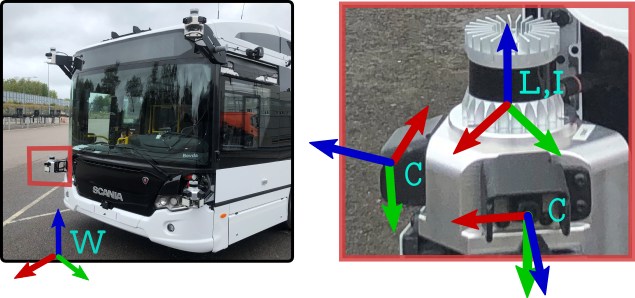} 
		\vspace{-7mm}
		\caption{Reference frames conventions for our vehicle platform. The world frame $\World$ is a fixed frame, while the base frame $\Base$, as shown in \Figure \ref{fig:scania-bevda}, is located at the rear axle center of the vehicle. Each sensor unit contains the two optical frames $\Camera$, an IMU frame, $\Imu$, and lidar frame $\Lidar$.} 
		\label{fig:coordinate-frames}
		\vspace{-4mm}
	\end{figure}
	We start by describing the necessary notation and reference frames used in our system according to the convention by Furgale~\cite{furgale2014representing}. The reference frames are shown in \Figure \ref{fig:coordinate-frames}. The vehicle base frame $\Base$ is located on the center of the rear-axle of the vehicle. Sensor readings from lidars, cameras and IMUs are represented in their respective sensor frames as $\Lidar_k$, $\Camera_k$ and $\Imu_k$ respectively, where $k$ denotes the location of the sensor in the vehicle. 

	For example, $\Lidar_{F}$ and $\Lidar_{R}$ represent the front right and rear left lidar frames. In our system we calibrated $\Lidar_{R}$ with respect to $\Lidar_{F}$ and used an known transformation between $\Base$ and $\Lidar_{F}$ from the vehicle CAD files. After calibration, we converted the extrinsics to vehicle base frame for verification and visualization purpose discussed in Sec. \ref{verification}.
	

	
	%
	
	\subsection{Pose Estimation Frames}
	For pose estimation of individual lidars we computed the relative transformation between subsequent lidar readings in its corresponding sensor frame. Conscutive estimates were available at times $t_{i-1}$ and $t_i$, denoted as ${\T}{_{\Lidar_{i-1}\Lidar_{i}}}$, where $\T = \left[\begin{array}{ll}\mathbf{R_{3\times3}} & \mathbf{t_{3\times1}} \\ \mathbf{0}^{\top} & 1\end{array}\right]\in \SEthree$ represents a homogeneous transformation consisting of rotation, $\mathbf{R} \in \SOthree$ and translation $\mathbf{t} \in \Real^3$. In our system we denote the iterative lidar pose transforms as ${\T}{_{F_{i-1}F_{i}}}$ and ${\T}{_{R_{i-1}R_{i}}}$. For brevity of discussion, we will denote them as ${\T}{_{F}}$ and ${\T}{_{R}}$ respectively. In future, we would extend this approach to solve the multi-lidar calibration scenario as the data collection vehicle is equipped with more than two lidars, as shown in Fig. \ref{fig:coordinate-frames}.

	\subsection{Problem Formulation}
	
	\begin{figure}[!t]
		\centering
		\vspace{2mm}
		\includegraphics[width=1\columnwidth]{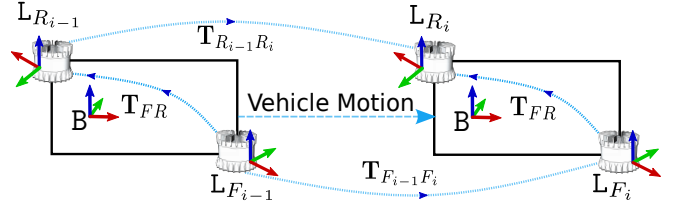}
		\caption{Transformation in between the lidars $\Lidar_{F}$ and $\Lidar_{R}$ in between times $t_{i-1}$ and $t_{i}$.}
		\label{fig:hand-eye}
		\vspace{-7mm}
	\end{figure}	
	
	Given a set of point clouds, $\calP_{\Lidar_{k:1(1)\calK}}^{t_{i:1(1)\calN}}$, from $\calK$ different lidars in a time sequential order $t_1 \dots t_\calN$, in their respective sensor frames, we wish to recover the extrinsics between the lidars as, ${\T}{_{\Lidar_{a}\Lidar_{b}}}$, where $a,b:1(1)\calK$ and $a\neq b$. We assume the lidars are fixed rigidly to the vehicle and all the other lidars are calibrated with respect to a reference lidar. The multi-lidar calibration task is formulated in terms of a Hand-eye \cite{Tsai_Hand_Eye_1989} calibration problem, where the extrinsics are recovered by matching the homogeneous transformations of the multiple lidars. For brevity of discussion we will consider two non-overlapping FoV lidars, the front right and rear left, and aim to recover the extrinsics. However, it can be extended to multi-lidar setup as well. With regards to the Hand-Eye method, we can recover the extrinsic calibration in between the lidars, as shown in \Figure \ref{fig:hand-eye}, by using the following equation,  
	
	\vspace{-7pt}
	\beq
	{\T}{_{F}}{\T}{_{FR}} = {\T}{_{FR}}{\T}{_{R}}
	\label{eq:hand-eye}
	\eeq
	
	which can be further decomposed into,
	\vspace{-3pt}
	\beq
	\begin{aligned}
		\mathbf{R}_{F}\mathbf{R}{_{FR}} &= \mathbf{R}{_{FR}}\mathbf{R}{_R} \\
		(\mathbf{R}{_F} - \Ithree) \mathbf{t}_{FR} &= \mathbf{R}{_{FR}} \mathbf{t}{_R} - \mathbf{t}{_F}.
		\label{eq:hand-eye-rt}
	\end{aligned}
	\eeq
	
	Multiple solutions including iterative and closed-form methods along with their benefits and drawbacks \cite{shah2012overview} have been discussed to solve the Hand-Eye calibration eq. \ref{eq:hand-eye} and \ref{eq:hand-eye-rt}. The solution methods can be further categorized as separable and simultaneous solutions. The separable solutions solve for $\mathbf{R_{3\times3}}$ and $\mathbf{t_{3\times1}}$ separately, whereas the simultaneous solutions solve them jointly. In our work, we have adapted a simultaneous approach using DQ \cite{Daniilidis1999}.
	
	\section{Methodology}
	\label{sec:methodology}
	
	\subsection{Initialization}
	Pose estimation of the individual sensors was computed using the VILENS estimator~\cite{wisth2020vilens}. However, instead of using a feature-based configuration, our configuration fused ICP-based point cloud matching~\cite{Pomerleau12comp} with IMU pre-integration using GTSAM \cite{Forster2017,Dellaert2017} to  estimate the motion of the individual lidars and to correct for motion distortion effects. We used the set of estimated poses $\mathbf{t_{F}}$ and $\mathbf{t_{R}}$ recovered at $\calN$ time iterations to estimate an initial ${\T}{_{FR}}$ by solving the following least-squares optimization,
	
	
	\begin{equation}
		\mathbf{R}^*, \mathbf{t}^*  = \argmin_{\mathbf{R}, \mathbf{t}} \|\mathbf{R}\mathbf{t}_{F} + \mathbf{t} - \mathbf{t}_{R}\|^2.
		\label{eq:kabash-alignment}
	\end{equation}  
	We used the closed form solution proposed by Umeyama \cite{umeyama1991least} to recover the initial optimal $\mathbf{R}^*$ and $\mathbf{t}^*$ which gives similar results as Horn et al. \cite{horn1987closed} in quaternion space. This initialization was necessary for the DQ based optimizer to generate better results.

	\subsection{Dual Quaternion Based Hand-Eye Calibration}
	Homogeneous transformation matrices are orthogonal and due to floating point errors, operations on them often result in matrices that need to be re-normalized. Although, this can be achieved using the Gram-Schmidt method, this is however slower than quaternion normalization. DQs are the natural extension to quaternions; because, unlike quaternions, they can handle translation as well in the same compact notion.
	
	\subsubsection{Preliminaries}
	We review DQ's and its operations briefly; more details are available here \cite[p.~53--62]{mccarthy1990introduction}. DQ has the advantages of a quaternion such as gimbal lock problem and unambiguous representation but with less intensive operations for transformations. A DQ, $\mathbf{q}$, can be written as,
	
	\begin{equation}
		\mathbf{q}=q_{r}+\varepsilon q_{d},
		\label{eq:dq}
	\end{equation} 
	
	\noindent where the real part $q_{r}$, and the dual part $q_{d}$, are quaternions. $\varepsilon$ is the dual unit, defined by $\varepsilon^2 = 0$ and $\varepsilon \ne 0$. The conjugate of a DQ, $\mathbf{q}$ is:
	\begin{equation}
		\mathbf{q^\star}=q_{r}^\star+\varepsilon q_{d}^\star,
		\label{eq:dq-conjugate}
	\end{equation} 
	where, $q^\star = (q_w,\text{-}q_x,\text{-}q_y,\text{-}q_z)$, is quaternion conjugate. Some operations of DQs, $\mathbf{q_1}$ and $\mathbf{q_2}$ are as follows:
	\begin{align}
		\mathbf{q_1} + \mathbf{q_2} &= (q_r^{1}+q_r^{2})+\varepsilon (q_d^{1} + q_d^{2}) \\
		\mathbf{q_1}  \otimes  \mathbf{q_2} &= q_r^{1} \otimes q_r^{2} +\varepsilon (q_r^{1} \otimes q_d^{2} + q_d^{1} \otimes q_r^{2}) \\
		\|\mathbf{q}\|^2 &= \mathbf{q}  \otimes  \mathbf{q}^\star \\
		\mathbf{q^{-1}} &= \mathbf{q}^\star, \quad \text{if } \|\mathbf{q}\|^2 = 1,
		\label{eq:dq-operations}
	\end{align}
	
	
	where, $\otimes$ and $\|\mathbf{\cdot}\|^2$ represent quaternion product and norm of a DQ respectively. 
	Given a transformation, $\T (\mathbf{R}, \mathbf{t})$ we can convert it to DQ form  as,
	\begin{equation}
		\mathbf{q(\T)}=q(\mathbf{R})+\varepsilon \frac{1}{2}q(\mathbf{t})q(\mathbf{R})
		\label{eq:T_to_dq}
	\end{equation} 
	where, $q(\mathbf{R})$ is obtained from Euler-Rodrigues formula and $q(\mathbf{t}) = (0,t_x,t_y,t_z)$. Conversely, we can convert a DQ, $\mathbf{q}(q_r, q_d)$ to a homogeneous transformation form as,
	
	\begin{equation}
		\T(\mathbf{q}) = \left[\begin{array}{cc}\mathbf{R}(q_{r}) & 2 q_{d} \otimes q_{r}^{*} \\ 0^{\top} & 1\end{array}\right]
		\label{eq:dq_to_T}
	\end{equation}
	where, $\mathbf{R}(q_{r})$ is obtained from Rodrigues' rotation formula.
	
	
	\subsubsection{Pose alignment}
	Since, the estimated poses are not necessarily time synchronized we aligned them using ScLERP \cite{kavan2005spherical}, which is a generalization of SLERP \cite{shoemake1985animating} before solving the Hand-Eye problem. The interpolation between two quaternions is given by SLERP whereas, the interpolation between two DQs is given by ScLERP as, 
	\beq
	\begin{aligned}
		\text{ScLERP}(n;\mathbf{q_1}, \mathbf{q_2}) &= \mathbf{q_1} \otimes (\mathbf{q_1}^{-1} \otimes \mathbf{q_2})^n \\
	\end{aligned}
	\eeq
	where, $n \in [0,1]$ and $\mathbf{q}^n = \cos(\frac{n\theta}{2}) + \mathbf{k} \cdot \sin(\frac{n\theta}{2})$. Here, $\theta$ is the dual angle of rotation around the unit DQ, $\mathbf{k}$ with zero scalar part. This ensures synchronized lidar poses at all timestamps, improving Hand-Eye calibration results.
	
	\subsubsection{Solving Hand-Eye Calibration}
	To solve the Hand-Eye calibration problem in DQ space, we initially converted the lidar transformations in between subsequent timestamps in DQ formulation. $\mathbf{q_{{\textstyle\mathstrut}F}}$ and $\mathbf{q_{{\textstyle\mathstrut}R}}$ represent the front right and rear left converted transformations in DQ space respectively, whereas, $\mathbf{q}$ represents the extrinsics in DQ formulation. Similar, to eq. \ref{eq:hand-eye}, the screw motion concatenation in DQ space yields, 
	\beq
	\mathbf{q_{{\textstyle\mathstrut}F}} = \mathbf{q} \mathbf{q_{{\textstyle\mathstrut}R}} \mathbf{q^\star}.
	\label{eq:dq_transformation}
	\eeq
	From Screw Congruence Theorem \cite{chen1991screw} it has been shown only the vector parts in the quaternion contribute to the derivation of unknown DQ, $\mathbf{q}$, as the scalar parts are equal. The vector part $\vv{q}$, in a quaternion $q$ is, $(0, \vv{q})$. With the derivations shown in \cite{Daniilidis1999}, eq. \ref{eq:dq_transformation} can be re-written as,
	
	\beq
	\resizebox{\columnwidth}{!}{$
		\left[\begin{array}{cccc}\vv{q_{r}^\text{F}}-\vv{q_{r}^\text{R}} & {[\vv{q_{r}^\text{F}}+\vv{q_{r}^\text{R}}]_{\times}} & \mathbf{0}_{3 \times 1} & \mathbf{0}_{3 \times 3} \\
			& & & \\	
			\vv{q_{d}^\text{F}}-\vv{q_{d}^\text{R}} & {[\vv{q_{d}^\text{F}}+\vv{q_{d}^\text{R}}]_{\times}} & \vv{q_{r}^\text{F}}-\vv{q_{r}^\text{R}} & {[\vv{q_{r}^\text{F}}+\vv{q_{r}^\text{R}}]_{\times}}\end{array}\right]\left[\begin{array}{l}q_r \\\\ q_d\end{array}\right]=0,$
		\label{eq:dq_T_matrix}}
	\eeq
	where, $[\:\cdot\:]_{\times}$ represents a skew-symmetric matrix and the first term, $\mathbf{S}$ is a 6$\times$8 matrix and the second term, is a 8$\times$1 vector. Based on the data collected at $\calN$ different timestamps we can construct a 6$\calN\times$8 matrix, $\mathbf{M} = (\mathbf{S}^\text{T}_1 \quad \mathbf{S}_2^\text{T} \quad \dots \quad \mathbf{S}^\text{T}_\calN)^\text{T}$. However, as this is not a full-rank matrix, the last two right singular vector obtained from SVD of $\mathbf{M}$ spans the nullspace of $\mathbf{M}$, which is used to recover $\mathbf{q}$. We also recorded transformations in non-planar trajectories as mentioned in \cite{Daniilidis1999} for recovering solutions of eq. \ref{eq:dq_T_matrix} and solved it within a certain batch size, instead of applying SVD on the whole 6$\calN\times$8 matrix at once.  
	
	\subsection{Calibration Verification}
	\label{verification}
	
\begin{figure} [!t]
		\centering
		\includegraphics[width=1\columnwidth]{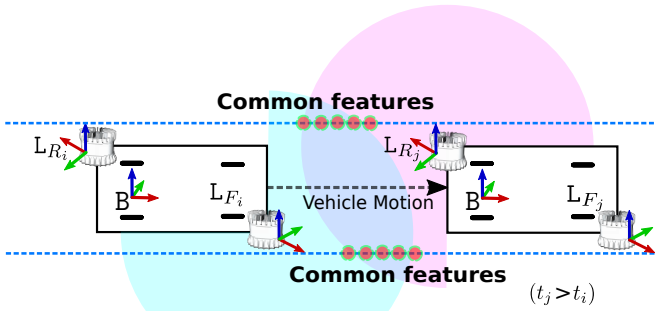}
		\vspace{-9mm}
		\caption{Semantic feature matching in between $L_F$ and $L_R$ lidars viewed at $\delta t = t_j-t_i$ time apart so that FoVs correspond to similar region. First, lidar $L_F$ detects a set of features at time $t_i$ and $\delta t$ time later $L_R$ observes the previously observed set of features (by $L_F$) at time $t_j$.}
		\vspace{-7mm}
		\label{fig:lidar-semantic-matching}
	\end{figure}

	To verify the accuracy of extrinsics, we associated extracted lidar semantic features between the lidars. The semantic features from lidar were extracted by associating the projected point clouds into the segmented camera images. The image segmentation was performed using a trained Efficient-Net model. We used an unsupervised method to calculate a time offset based on vehicle kinematics which described the observation delay between the lidars, corresponding to similar features as represented in Fig. \ref{fig:lidar-semantic-matching}.
	
	\subsubsection{Efficient-Net for semantic segmentation}
	
	
	The author of EfficientNet \cite{tan2019efficientnet} demonstrated that model scaling can be achieved by carefully balancing network depth, width, and resolution, leading to a better performance with the fixed computation resources. Based on this study, we used a modified version of EfficientNet to semantically segment the camera images. The modified network architecture backbone is shown in Table \ref{tab:efficientNet}, where each row describes a stage $i$ with $\hat{L}_{i}$ layers, with input resolution $\langle\hat{H}_{i}, \hat{W}_{i}\rangle $ and output channels $\hat{C}_{i}$. MBConv layers are mobile inverted bottlenecks from the MobileNetV2 architecture \cite{sandler2018mobilenetv2} where squeeze-and-excitation optimization have also been added from \cite{hu2018squeeze}. The segmentation head consists of convolutional layers and bilinear interpolation to upsample the network output to its original input resolution.
	
	\begin{table}[!h]
		\normalsize
		\centering
		\vspace{-2mm}
		\resizebox{\columnwidth}{!}{
			\begin{tabular}{c|cccc} \toprule
				Stage & Operator & Resolution & \#Channels & \#Layers \\
				\( i \) & \( \hat{\mathcal{F}}_{i} \) & \( \hat{H}_{i} \times \hat{W}_{i} \) & \( \hat{C}_{i} \) & \( \hat{L}_{i} \) \\
				\midrule
				1 & Conv3x3 & \( 640 \times 1024 \) & 32 & 1 \\
				2 & MBConv1, k3x3 & \( 320 \times 512 \) & 16 & 1 \\
				3 & MBConv6, k3x3 & \( 160 \times 256 \) & 24 & 2 \\
				4 & MBConv6, k5x5 & \( 80 \times 128 \) & 40 & 2 \\
				5 & MBConv6, k3x3 & \( 40 \times 64 \) & 80 & 3 \\
				6 & MBConv6, k5x5 & \( 40 \times 64 \) & 112 & 3 \\
				7 & MBConv6, k5x5 & \( 20 \times 32 \) & 192 & 4 \\
				8 & MBConv6, k3x3 & \( 20 \times 32 \) & 320 & 1 \\
				9 & Segmentation Head & \( 640 \times 1024 \) & 24 & - \\
				\bottomrule
		\end{tabular}}
		\caption{ Modified EfficientNet-B0 baseline network architecture.}
		\label{tab:efficientNet}
		\vspace{-6mm}
	\end{table}
	
	
	\subsubsection{Association of semantic features}
	We chose to match curb features between the lidars because the lidar projection error was stable across frames for the curb semantics. First, we sampled an arbitrary number of curb semantic points from the $L_F$ lidar by projecting the points into camera perspective, for a specific duration to extract a segment of the curb. Then based on the calculated time offset obtained from vehicle kinematics, we sampled curb semantics from the $L_R$ lidar. Since the point clouds of both lidars were represented in the base frame, we expected that the same set of features should be observable from both lidars. Finally we calculated the root mean square error (RMSE) between the commonly observed curb features. The association between the feature points was established using tri-linear interpolation on the sorted points. We therefore sampled several curb features using the aforementioned technique and statistically compared RMSE before and after the calibration. Curb extraction and association results are presented in Sec. \ref{semantic_verification}.

	
	
	\section{Experimental Results}
	\label{sec:results}
	
	\subsection{Dataset}
	We collected the data from a Scania autonomous bus platform consisting of 4 corner lidars and 8 cameras. For ground truth pose estimates (GT) we used data from a Novatel GNSS. All the sensing data was synchronized using PTP synchronization.
	
	\subsection{Results}	
	\subsubsection{Pose Estimation and DQ based Hand-Eye calibration}
\begin{figure} [!htbp]
		\centering
		\includegraphics[width=1\columnwidth]{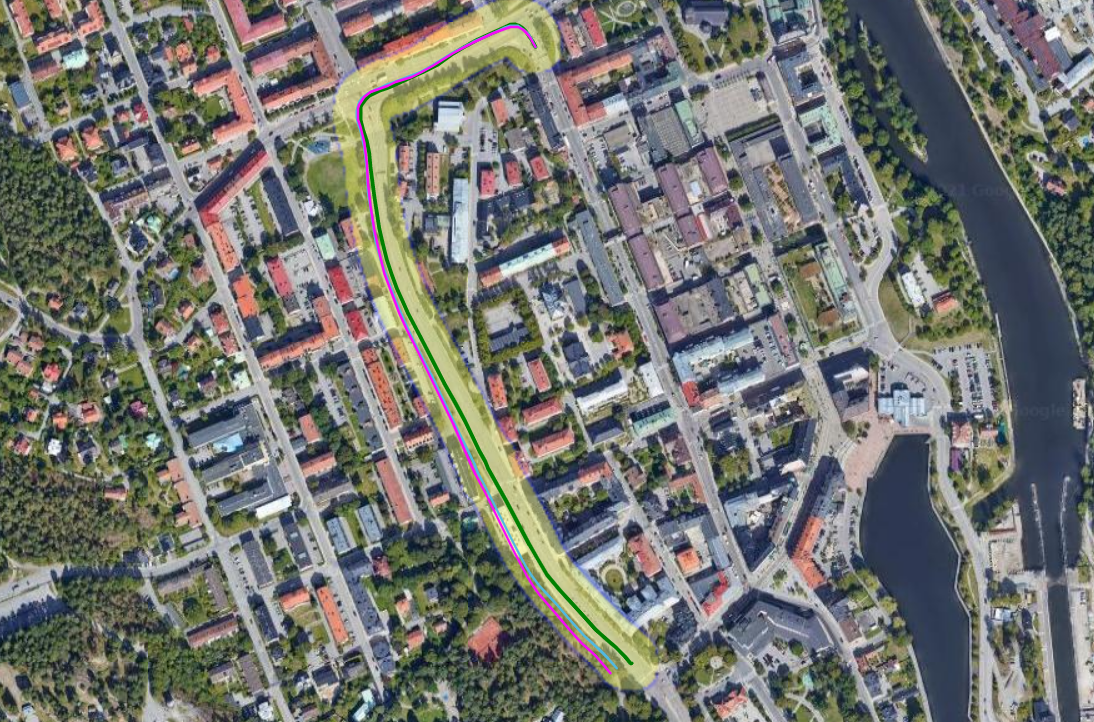}
		\caption{Results of the odometry estimation from $L_F$ (blue) and $L_R$ (magenta) lidars plotted in Google Maps with respect to the GT (green) with initial calibration parameters. The GT was generated using Novatel GNSS.}
		\label{fig:maps-odom}
	\end{figure}

\begin{figure} [!htbp]
		\vspace{-4mm}
		\centering
		\includegraphics[width=1.\columnwidth]{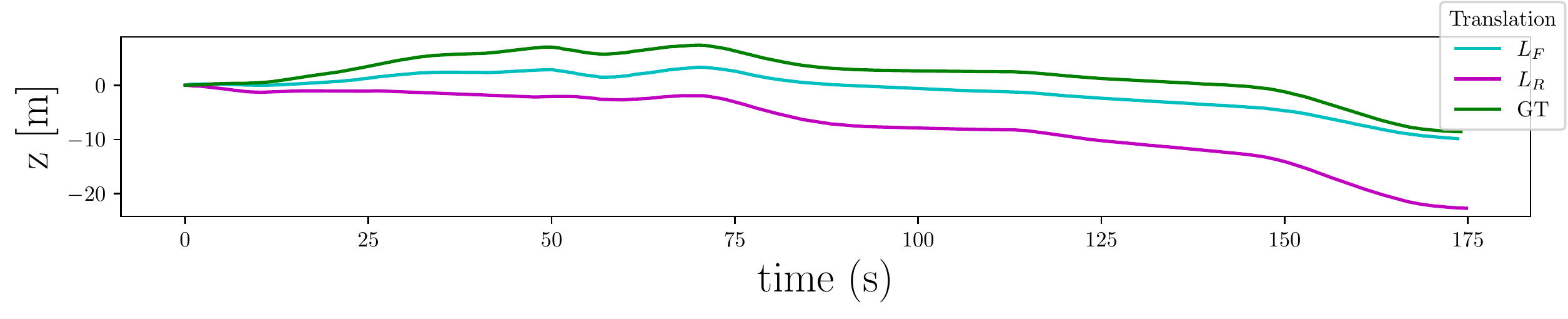}
		\vspace{0.01\columnwidth}
		\includegraphics[width=1.\columnwidth]{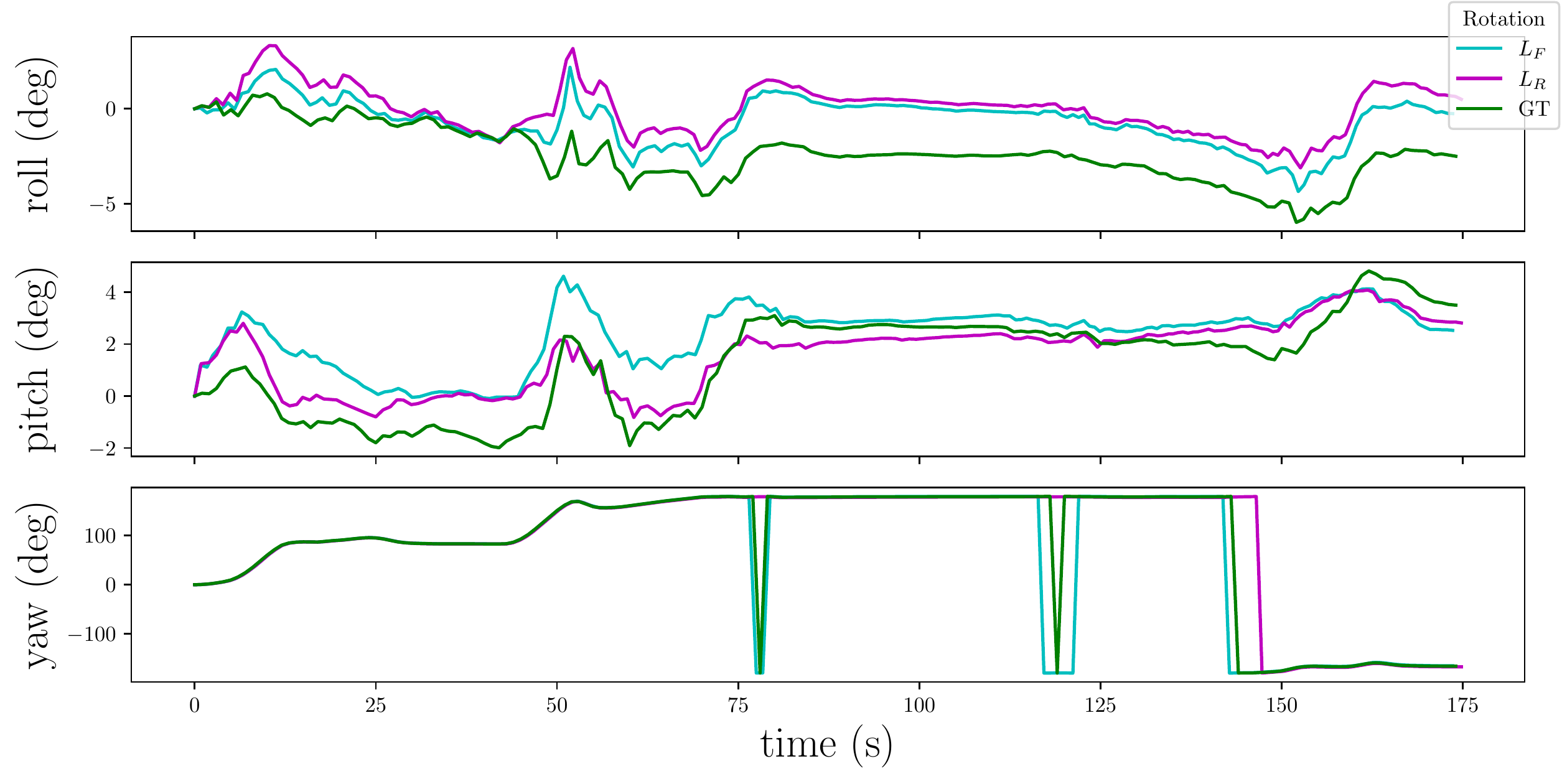}
		\vspace{-5mm}
		\caption{Pose estimation for $L_F$ and $L_R$ lidars with respect to GT with initial calibration estimation.}
		\label{fig:init-calibration}
		\vspace{-1mm}
	\end{figure}

	We estimate the poses of the $L_F$ and $L_R$ lidars and convert them to the vehicle base frame for comparison. The pose estimates are shown in Fig. \ref{fig:maps-odom} and Fig. \ref{fig:init-calibration} along with the GT for the initial calibration parameters from the vehicle model. 
	
	The relative pose error (RPE) and absolute pose error (APE) metrics are used to evaluate the local and global consistency respectively of the estimated trajectory of $L_R$ lidar with respect to $L_F$ lidar, before and after calibration with the Hand-Eye method after time alignment. The ScLERP based time alignment increased the consistency in between the DQ poses for the scalar part match which was a pre-condition for deriving eq. \ref{eq:dq_T_matrix}. We re-estimate the poses of the $L_R$ lidar after putting in new calibration parameters. 
	
	RPE compares the difference between the individual relative poses within two timestamps $t_i$ and $t_j$ as, $RPE_{i,j} = ({\T}{_{\Lidar_{F}(t_i)}^{-1}}{\T}{_{\Lidar_{F}(t_j)}})^{-1}({\T}{_{\Lidar_{R}(t_i)}^{-1}}{\T}{_{\Lidar_{R}(t_j)}})$. APE is based on the absolute relative pose between two poses ${\T}{_{\Lidar_{F}(t_i)}}, {\T}{_{\Lidar_{R}(t_i)}} \in \SEthree$ at timestamp $t_i$ as, $APE_{i} = {\T}{_{\Lidar_{F}(t_i)}^{-1}}{\T}{_{\Lidar_{R}(t_i)}}$.	
	\begin{figure} [h!]
		\vspace{-3mm}
		\centering
		\includegraphics[width=1.\columnwidth]{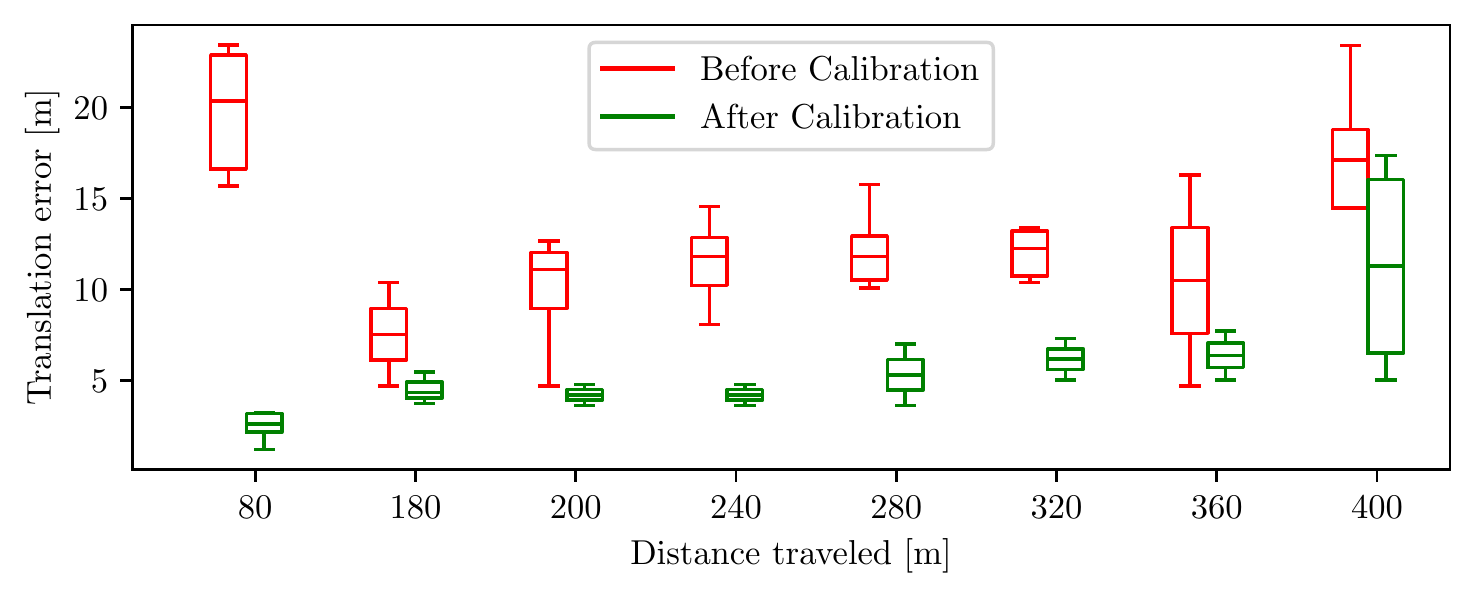}
		\includegraphics[width=1.\columnwidth]{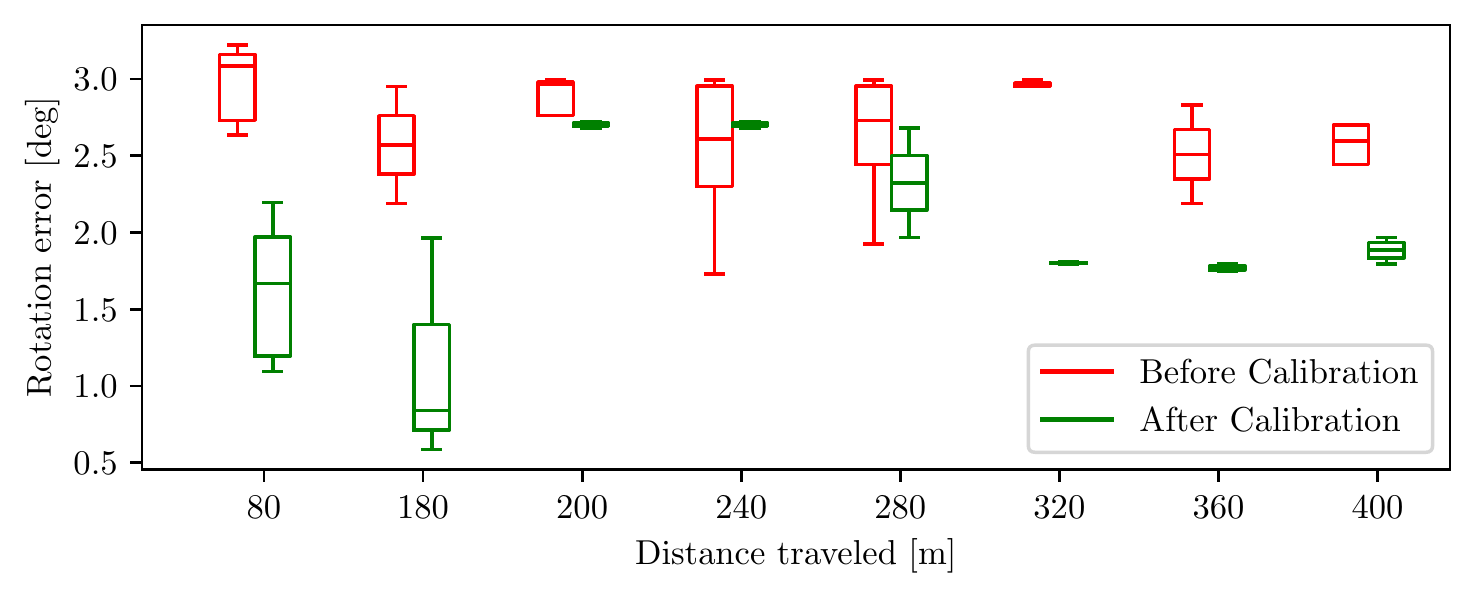}
		\vspace{-5mm}
		\caption{RPE between $L_F$ and $L_R$ lidars before and after calibration.}
		\label{fig:calibration-rpe}
		\vspace{-2mm}
	\end{figure}
	\begin{figure} [h]
		\vspace{-7mm}
		\centering
		\includegraphics[width=1.\columnwidth]{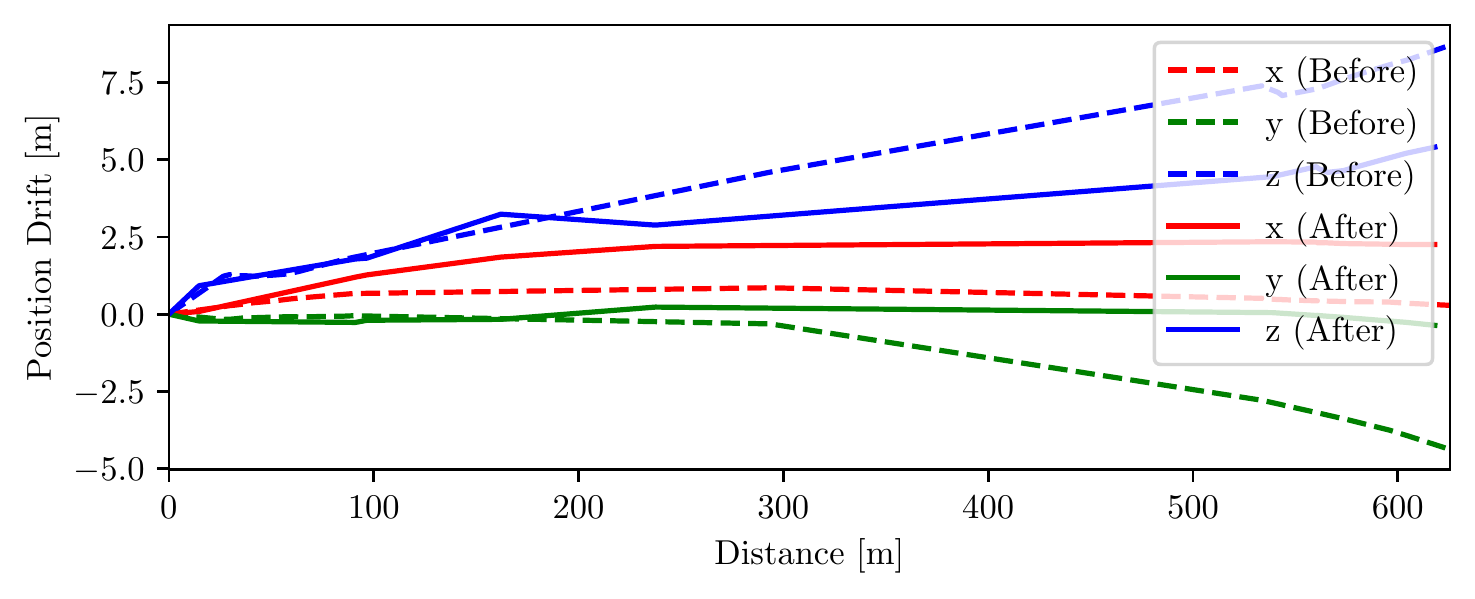}
		\includegraphics[width=1.\columnwidth]{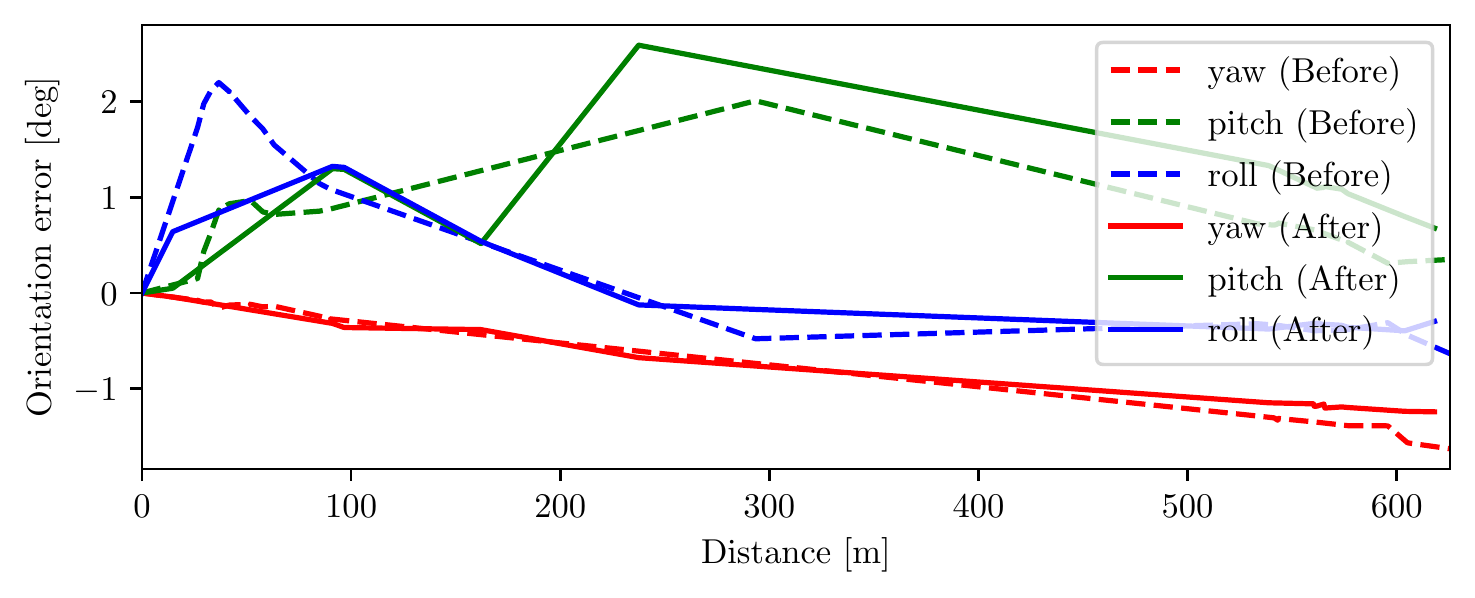}
		\vspace{-8mm}
		\caption{APE between $L_F$ and $L_R$ lidars before and after calibration.}
		\label{fig:calibration-ape}
	\end{figure}
	We compare the RPE, Fig. \ref{fig:calibration-rpe} and APE, Fig. \ref{fig:calibration-ape} between $L_F$ and $L_R$ lidar trajectories before and after calibration, which shows that we recover better calibration estimates in between the $L_F$ and $L_R$ lidar.

	\subsubsection{Semantic verification}
	\label{semantic_verification}
	For semantic verification we train the EfficientNet model on a diversified scenario set (city driving, snowy conditions, dessert area, suburban area). The overall training accuracy is 79.10\% and the mean IoU (Intersection over Union) is 0.495. Whereas, for curb class the accuracy is 67.20\% and the IoU is 0.557. The model performance is correlated with the amount of data it is trained on. We extract the curb points by doing a perspective projection of the lidar points in image space and choosing points closer to the curb pixels within a bound of $\pm$5 pixels.

	\begin{figure} [!htbp]
		\vspace{1mm}
		\centering
		\includegraphics[width=1\columnwidth]{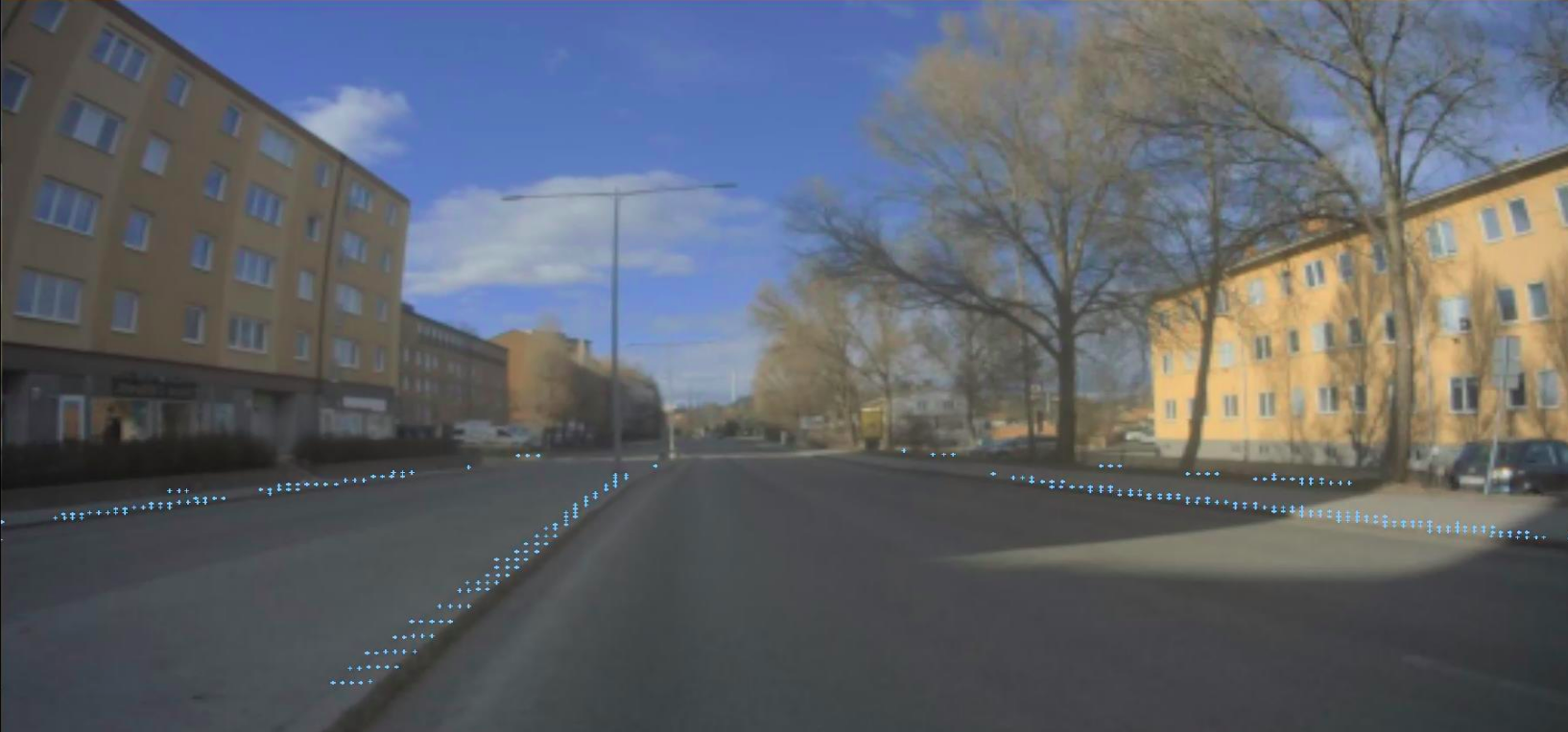}
		\vspace{-7mm}
		\caption{Lidar point cloud projected into camera perspective with filtered curb semantics, shown as cyan colored points.}
		\vspace{-4mm}
		\label{fig:pc-projection}
	\end{figure}

	
	\begin{figure} [!htbp]
		\vspace{1mm}
		\centering
		\includegraphics[width=1\columnwidth]{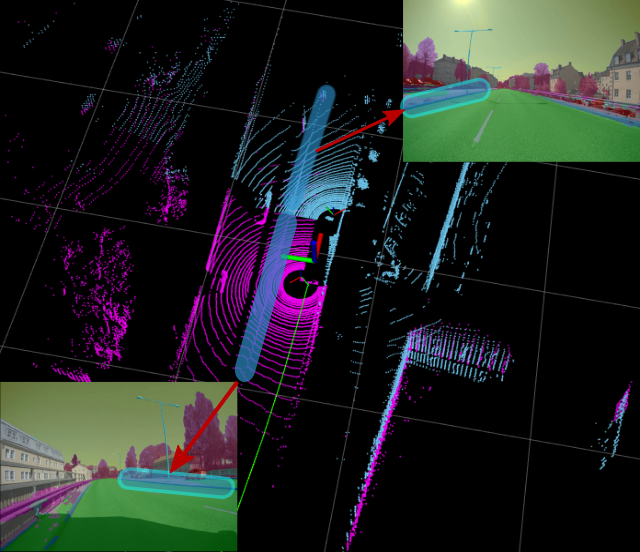}
		\vspace{-7mm}
		\caption{Calibrated lidar point clouds from $L_F$ and $L_R$ lidar with their respective image visuals with corresponding curb features highlighted.}
		\vspace{-5mm}
		\label{fig:dnn-extrapolation}
	\end{figure}
	As seen in Fig. \ref{fig:pc-projection}, we extract the curb points from the front lidar. And similar extraction was performed on the rear lidar as well. After calculating the time offset based on vehicle kinematics we could verify for the calibrated scenario the RMSE was statistically lower. A full representation of the calibrated lidar point clouds are shown in Fig. \ref{fig:dnn-extrapolation}.
	
	\section{Conclusion}
	\label{sec:conclusion}
	We have presented a multi-lidar calibration method that 
	helps in automatically determine and verify the quality of the extrinsic calibration parameters between the lidars. We have also shown that aligning the poses with ScLERP and applying the DQ based Hand-Eye calibration improves the quality of the extrinsic calibration parameters. In future, the extrinsics can be used in state vector in SLAM solutions and be recovered online by solving the DQ based Hand-Eye calibration. Furthermore, the verification part of the extrinsics can be improved by calculating the correlation in between the features of individual lidars in a common map reference frame. For the verification purpose we assumed the perspective projection from the lidar to camera did not introduce any errors into the system. This can be improved by extracting semantic features directly from the point cloud by training a supervised model on labeled point cloud data.

	\section{Acknowledgements}
	This research has been conducted as part of development of autonomous transport solutions at Scania, Sweden. It was jointly funded by Swedish Foundation for Strategic Research (SSF) and Scania. The research was also affiliated with WASP.
	
	
	\bibliographystyle{./IEEEtran}
	\bibliography{./IEEEabrv, ./library}
	
	
\end{document}